\documentclass{article}

\usepackage{microtype}
\usepackage{graphicx}
\usepackage{subfigure}
\usepackage{booktabs} 

\usepackage{hyperref}

\usepackage[accepted]{icml2025}
\usepackage{amsmath}
\usepackage{amssymb}
\usepackage{mathtools}
\usepackage{amsthm}

\usepackage[capitalize,noabbrev]{cleveref}

\usepackage{lipsum}         
\usepackage{amsmath}         
\usepackage{enumitem}
\usepackage{float}
\usepackage{multirow}

\setlist[enumerate]{topsep=1pt, partopsep=1pt}
\theoremstyle{plain}

\theoremstyle{definition}

\theoremstyle{remark}

\def \title {Koopman Autoencoders Learn Neural Representation Dynamics}

\icmltitlerunning{\title}
\begin{document}

\twocolumn[
\icmltitle{\title}



\icmlsetsymbol{equal}{*}

\begin{icmlauthorlist}
\icmlauthor{Nishant Suresh Aswani}{tandon,nyuad}
\icmlauthor{Saif Eddin Jabari}{tandon,nyuad}
\end{icmlauthorlist}

\icmlaffiliation{tandon}{New York University Tandon, Brooklyn, USA}
\icmlaffiliation{nyuad}{New York University Abu Dhabi, Abu Dhabi, UAE}

\icmlcorrespondingauthor{}{nishantaswani@nyu.edu}

\icmlkeywords{Machine Learning, ICML}

\vskip 0.3in
]



\printAffiliationsAndNotice{}  

\begin{abstract}
This paper explores a simple question: can we model the internal transformations of a neural network using dynamical systems theory? We introduce Koopman autoencoders to capture how neural representations evolve through network layers, treating these representations as states in a dynamical system. Our approach learns a surrogate model that predicts how neural representations transform from input to output, with two key advantages. First, by way of lifting the original states via an autoencoder, it operates in a linear space, making editing the dynamics straightforward. Second, it preserves the topologies of the original representations by regularizing the autoencoding objective. We demonstrate that these surrogate models naturally replicate the progressive topological simplification observed in neural networks. As a practical application, we show how our approach enables targeted class unlearning in the Yin-Yang and MNIST classification tasks.
\end{abstract}

\section{Introduction}

Neural networks are defined by compositions. At each step, they transform their inputs, increasing the complexity of the overall transformation applied to data. Remarkably, these transformations have the effect of producing simple shapes at the output \cite{papyan2020}, when quantified by topology \cite{naitzat2020}. In fact, the neural representations (i.e., outputs of intermediate layers) of a network progressively simplify until a network arrives at the final output. This progression, along with the compositional nature of these networks, inspires an intuitive `path' perspective \cite{lange23a}. In other words, there is a notion of `traveling' some distance from the input to the output, along the path defined by these neural representations. Our work further explores this path analogy by asking: 

\begin{figure}[t]
\begin{center}
    \centerline{\includegraphics[width=0.95\columnwidth]{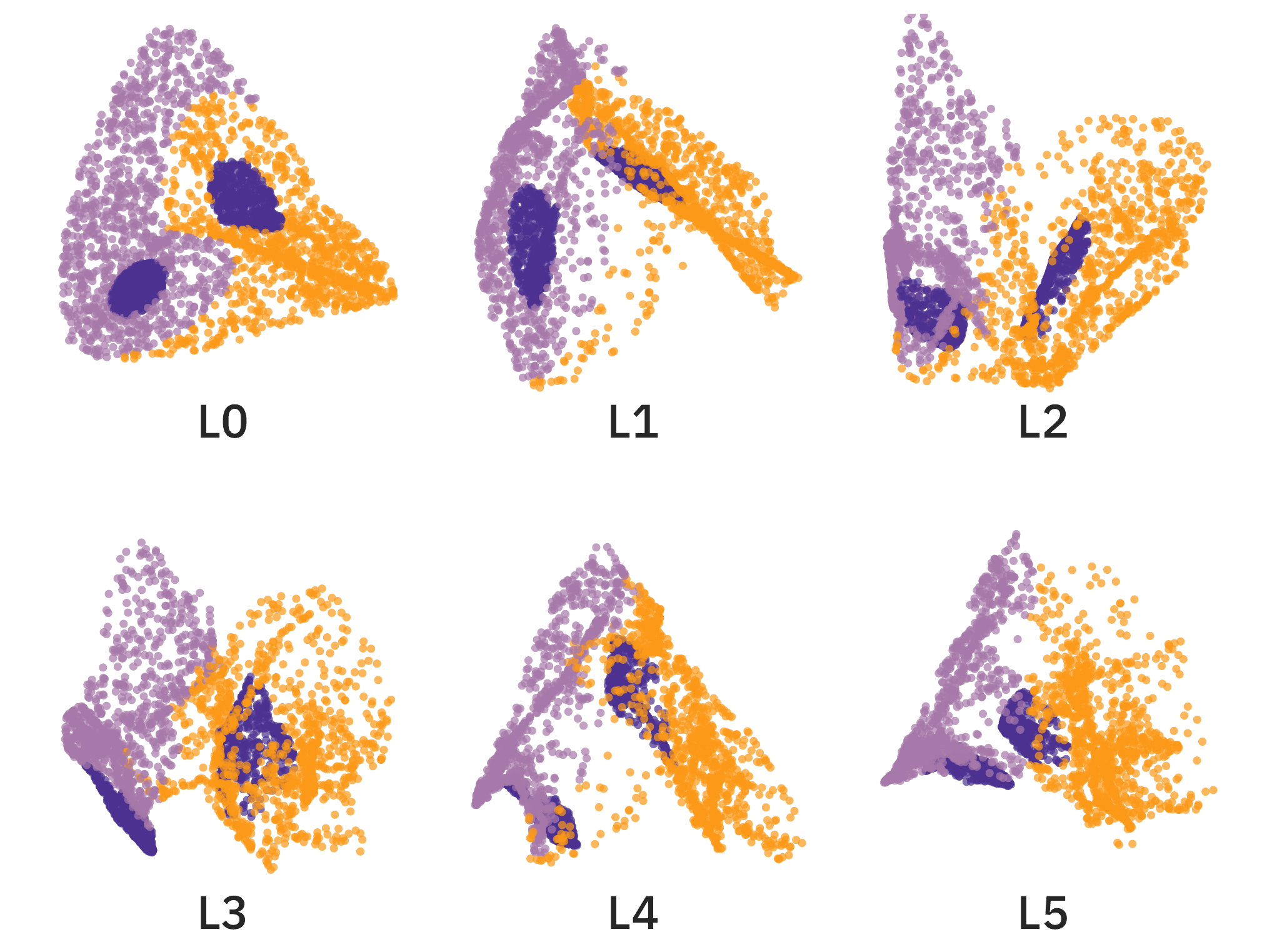}}
    \caption{The top three principal components of the neural representations from the first layer (L0) and all residual blocks (L1-5) of a multi-layer perceptron (MLP) with a ResNet-style architecture. Each plot contains $2\times10^3$ points and undergoes the preprocessing steps outlined in Section \ref{sec:preprocessing} before PCA for plotting. The model is trained on the Yin-Yang dataset \cite{kriener2022}, a three-way classification task. See Appendix \ref{app:dataset} for details on architecture and dataset.}
    \label{fig:yin_yang_activations}
\end{center}
\vskip -0.4in
\end{figure}

\textit{Can we discover a dynamics that generates this path?}

And, when equipped with the dynamics, we press on, exploring: 

\textit{Can we edit these dynamics to produce a different output than what was originally intended?}

To elaborate on the significance of our second question: editing, updating, or unlearning specific knowledge contained within neural networks prevents expensive retraining or removes harmful undesired outputs for model alignment \cite{yaoediting2023, gupta2024}. If these unwanted outputs lie at the end of our neural representation paths, then editing the dynamics can help us `steer away' from them, generating representations without these outputs.

Our work relies on modern Koopman-based approaches \cite{koopman1931hamiltonian, brunton2021modern, takeishi2017}. We learn our dynamics in an \textit{observable space}, different from the original space, defined by the latent space of a predictive, Koopman autoencoder. In observable space, our dynamics are defined by a linear operator, making the dynamics a simple object to work with.

\textbf{Contributions. } Our main contributions are as follows: 
\begin{itemize}[topsep=1pt, partopsep=1pt, itemsep=1pt, parsep=0pt]
    \item We introduce Koopman autoencoder surrogates as a framework for interpolating and editing the neural representations of a trained neural network. Our Koopman autoencoders generate realistic dynamics, producing intermediate outputs which follow our established understanding of how neural representations topologically simplify as they progress through the layers of a neural network.
    \item We develop an encoder isometry objective to supplement the optimization process of Koopman autoencoders, preserving the original topology of neural representations in observable space. 
    \item We demonstrate how our Koopman autoencoders can be used to edit neural representations in observable space, leading to fast, targeted class unlearning.
\end{itemize}

Overall, enabled by modern Koopman theory, our work develops a methodology to interpolate the neural representations of deep networks.

\begin{figure*}[ht!]
  \centering \includegraphics[width=\textwidth]{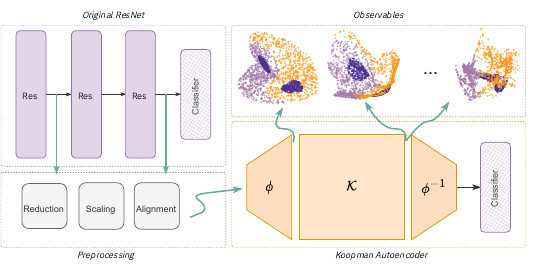}
  \vskip -0.1in
  \caption{A summary of our framework presented in Section \ref{sec:framework}. We gather neural representations from a trained, residual network and preprocess them to bring them into the same space. Afterwards, we train a Koopman autoencoder on a pair of the representations, resulting in predictive autoencoder with manipulable and visualizable observabe space.}
  \label{fig:hero}
\vskip -0.1in
\end{figure*}

\section{Preliminaries}

\subsection{Related Work}

\textbf{Topology and dynamics. } Our work is most closely aligned with literature that highlights topological and geometric perspectives in deep learning. Primarily inspired by \citet{naitzat2020}, we demonstrate how the shape of a data manifold can transform as it is processed by the layers of a neural network (NN). As advanced by \citet{lange23a}, we envision the outputs of each NN layer as forming a `path', arising naturally from the compositional structure of NNs. Additionally, we put to work an established dynamics perspective in deep learning. With a spotlight on deep residual networks (ResNets) \cite{he2016resnet}, there is growing evidence \cite{gai2021mathematical, li2023residual} that treats ResNet activations as traveling on a `conveyor belt' to their final output. This dynamics view plays nicely with the topological vantage, with \citet{naitzat2020} positing that ``[network] depth plays the role of time,'' in the sense that additional layers ``afford additional time to transform the data.'' 

Moreover, work on neural collapse \cite{papyan2020} has revealed how penultimate layer representations organize representations into a particular (simplex ETF) topology, which has spurred new training approaches \cite{markou2024}, exemplifying the value of understanding NN topology.

 

\textbf{Koopman-based approaches. } At the heart of our method is a Koopman autoencoder (KAE). KAEs have been employed in machine learning problems to forecast physical systems \cite{takeishi2017, lusch2018deep, azencot2020forecasting}, disentangle latent factors in sequential datasets \cite{berman2023multifactor}, and generate time-series \cite{naiman2023generative}. Traditionally, Koopman approaches find application in control tasks due to their predictive nature. Generally, practical approaches \cite{budivsic2012applied, brunton2021modern}, developed atop Koopman theory \cite{koopman1931hamiltonian}, work within a latent space equipped with linear dynamics allowing one to study, and potentially shape, these dynamics via linear control and spectral tools. Our work is unique in proposing a KAE to interpolate between and manipulate the topology of neural representations.

\textbf{Representation metrics.} Pertinent to our work are tools from representational similarity analysis (RSA) literature. Notably, \citet{kornblith19} discusses the required invariance properties of `dissimilarity' when comparing representations between neural network layers, with \citet{williams2021} extending these ideas to develop proper metrics. In maturing the `path' analogy, our work follows \citet{lange23a} by using tools from RSA to support our methodology. While our work does not require computing similarity metrics across representations, it does reason about the dynamics between them, demanding similar methodological care.

\textbf{Model editing.} As an application of our Koopman framework, we edit the linear operator which governs our dynamics. To achieve this, we use the EMMET algorithm \cite{gupta2024}, originally designed to update the weights in transformer blocks. As human knowledge and facts update, the field of model editing is concerned with updating large language models while avoiding expensive model retraining \cite{yaoediting2023}. While our work does not directly explore language models, we hypothesize that our framework can be extended to include the relevant architectures.

\subsection{Topology}

Our concrete measure of an object's topology refers to its \textit{Betti numbers}. For a $k$-dimensional manifold, one can compute $k$ Betti numbers, defining its topological signature. The zero-th Betti number, $\beta_0$, of a manifold refers to the number of unconnected components. The $k$-th Betti number, for $k\geq1$, quantifies the number of $k$-dimensional holes in the manifold. This manifests in the popular, though counterintuitive, quip that `a donut is topologically equivalent to a coffee mug.' Both objects have one connected component, a single 1-D hole, and zero 2-D holes, giving them the Betti number sequence $\beta=\{1,1,0\}$.

When working with discrete manifolds, such as neural representations from a network, quantifying topology relies on \textit{persistence homology}. Very simply, the approach computes $k$-dimensional \textit{simplices} (e.g., points, lines, triangles, tetrahedra, etc.) of an object at varying scales, which determine an object's \textit{homologies}. These homologies are closely related to the Betti numbers; by tracking these homology groups across scales, one can make claims about an object's topology. We rely on the \textit{Vietoris-Rips (VR) complex}, a particular method of computing the simplices, which in turn informs the Betti numbers. The VR complex requires a distance metric (in our case Euclidean) and a scale parameter $\epsilon$. For a more detailed background on algebraic topology we refer to \citet{naitzat2020}.

\subsection{Koopman theory}
In a typical discrete dynamical system, we observe measurements of a state $\mathbf{x}_t \in \mathcal{M} \subseteq \mathbb{R}^N$ at time $t \in \mathbb{Z^+}$, which evolve under a mapping $\mathcal{T: M \rightarrow M}$, such that 
\begin{equation}
\mathbf{x}_{k+1} = \mathcal{T}(\mathbf{x}_k).
\end{equation}
When $\mathcal{T}$ is nonlinear, these systems are often analyzed using linear approximations near fixed points, often to control the underlying nonlinear system.

Koopman operator theory suggests an alternative global linearization of the dynamics by finding a map in the \textit{observable space}, $\phi(x_k): \mathcal{M} \rightarrow \mathcal{F} \subseteq \mathbb{C}$. In this space, the linear map $\mathcal{K: F \rightarrow F}$, which evolves the observables, is defined as the \textit{Koopman operator.} If we assume our observables as vectors, we obtain the form
\begin{align}
\phi(\mathbf{x}_{k+1}) = \mathcal{K}\circ\phi(\mathbf{x}_k),
\end{align} 
where $\phi$ ``lifts" our original system states into the observable space resulting in a system that evolves under a linear operator. The forecast can be obtained in the state space by applying an inverse operation $\phi^{-1}: \mathcal{F} \rightarrow \mathcal{M}$ to the result of the forward dynamic. \citet{brunton2021modern} provide a fuller view of modern applications of Koopman theory, along with its rich history in machine learning. 

\begin{figure*}[ht!]
  \centering \includegraphics[width=\textwidth]{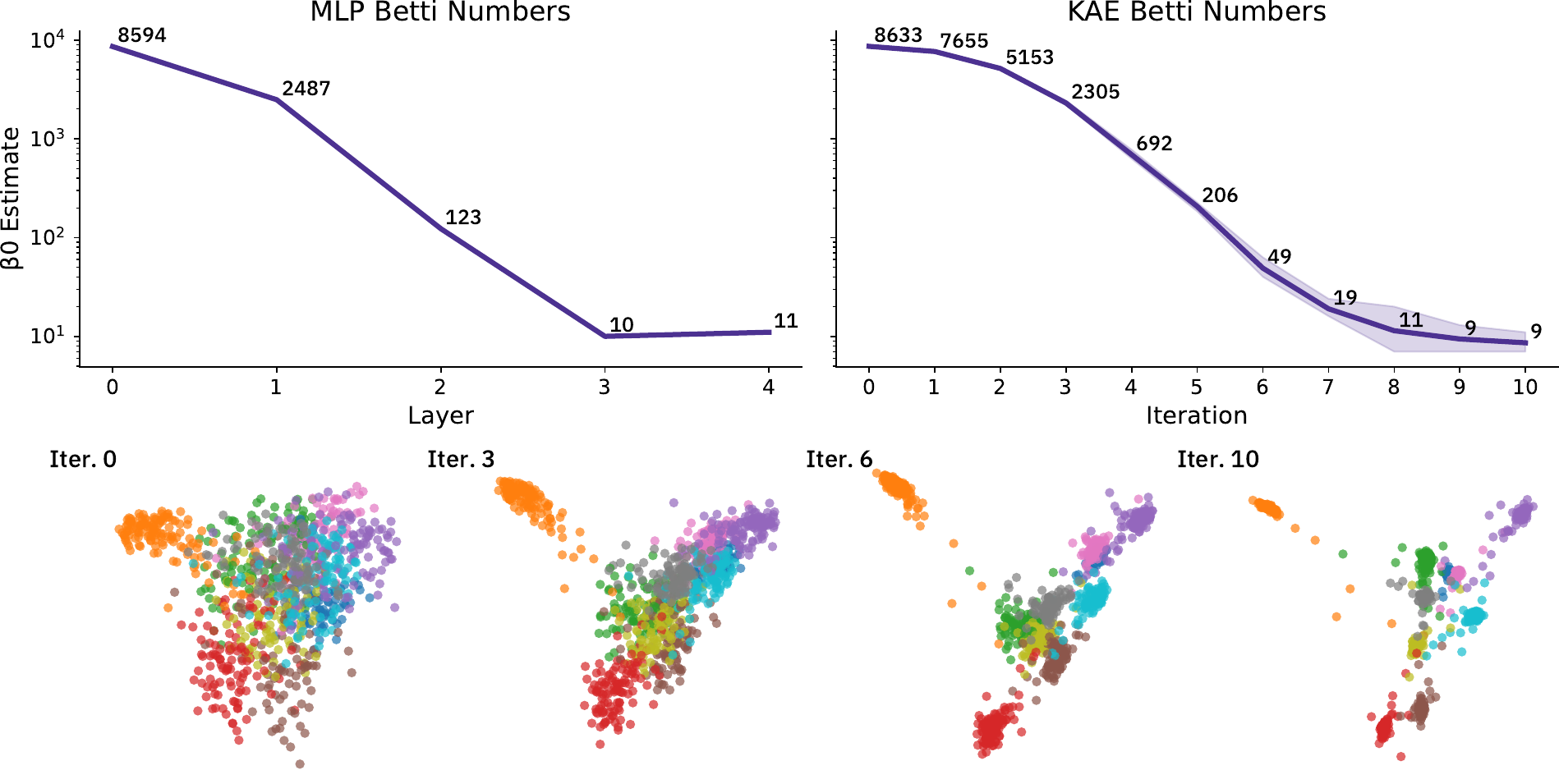}
  \vskip -0.1in
  \caption{(Top left) The $\beta_0$ Betti numbers of the neural representations from each residual block of a residual MLP trained on MNIST. The Betti numbers are computed using the Vietoris-Rips complex at a filtration $\epsilon=0.166$. (Top right) The average $\beta_0$ Betti numbers of intermediate outputs, projected into state space, for five KAEs trained on the first and penultimate layer representations of the residual MLP. The Betti numbers are computed using the Vietoris-Rips complex at a filtration $\epsilon=0.14$. (Bottom) Select intermediate outputs from an MNIST KAE, projected into the state space. At each successive iteration, the topology is simplified until it arrives at the penultimate layer representations.}
  \label{fig:topology}
\vskip -0.1in
\end{figure*}

\section{Koopman Autoencoders as Surrogates}
\label{sec:framework}
\subsection{Architecture}
Consider a trained neural network $\mathcal{N}^L$ composed of $L\in \mathbb{Z^+}$ layers, where each layer $f_i$ is indexed by $i \in \{1,2,...,L\}$. The network is defined by successive compositions, giving rise to the form
\begin{align}
\mathcal{N}^L(x) = f_{L} \circ \dots f_2\circ f_1(\mathbf{x}_0),    
\end{align}
where $\mathbf{x}_0$ is an input. The output of $f_i$ is the $i$-th neural representation $\mathbf{x}_{i} \in \mathbb{R}^{d_{i+1}}$, where $d_{i+1}$ is the input dimension of the subsequent layer $f_{i+1}$. Inspired by \citet{li2023residual}, we work with deep multi-layer perceptrons (MLPs) comprised of residual blocks, a form of residual networks (ResNets). Figure \ref{fig:yin_yang_activations} plots the top three principal components of neural representations from each residual block, visualizing how the data transform across the layers of a residual network.

We evoke a dynamical systems perspective of these ResNets, treating the neural representations $\{\mathbf{x}_1, \mathbf{x}_2, \dots, \mathbf{x}_L\}$ of the trained network as the states generated by a complex, nonlinear system. Within this context, we introduce a Koopman autoencoder, consisting of an encoder $\phi: \mathbb{R}^{d_{i+1}} \rightarrow \mathbb{R}^p$, a decoder $\phi^{-1}: \mathbb{R}^{p} \rightarrow \mathbb{R}^{d_{i+1}}$, and a linear operator $\mathcal{K}: p \rightarrow p$. In concert, they operate as
\begin{align}
\mathbf{x}_{j} = \phi^{-1}\circ\mathcal{K}\circ\phi(\mathbf{x}_i), \; \forall\, i,j \in \{1,2,\ldots,L\} : i < j
\label{eq:autoencoder}
\end{align}
In Equation \ref{eq:autoencoder}, $\phi$ embeds a neural representation into a (typically) higher-dimensional observable, after which $\mathcal{K}$ `advances' the observable. Finally, $\phi^{-1}$ returns the observable to the state space. We implement $\phi$ and $\phi^{-1}$ as symmetric, but untied, MLPs and define $\mathcal{K}$ as a learnable square matrix. Hence, the KAE produces a dynamic in the observable space, governed by the linear operator.

\subsection{Objectives}
The KAE is optimized with the objective functions
\begin{align}
\mathcal{L}_{\text{recon}} &= \left\| \mathbf{x}_{\{i,j\}} - \phi^{-1} \circ \phi(\mathbf{x}_{\{i,j\}}) \right\|^{2},
\label{eq:reconstruction}
\\
\mathcal{L}_{\text{linear}} &= \left\| \phi(\mathbf{x}_{j}) - \mathcal{K} \circ \phi(\mathbf{x}_{i}) \right\|^{2},
\label{eq:linear_prediction}
\\
\mathcal{L}_{\text{state}}  &= \left\| \mathbf{x}_{j} - \phi^{-1} \circ \mathcal{K} \circ \phi(\mathbf{x}_{i}) \right\|^{2},
\label{eq:state_prediction}
\\
\mathcal{L}_{\text{dist}}   &= \left\| \left\|\mathbf{x}_{\{i,j\}}\right\|^2 - \left\| \phi(\mathbf{x}_{\{i,j\}}) \right\|^2 \right\|^{2},
\label{eq:isometry}
\end{align}
resulting in a combined loss
\begin{align}
\mathcal{L_\text{total}} = \lambda_1 \mathcal{L}_{\text{recon}} + \lambda_2\mathcal{L}_{\text{linear}}+ \lambda_3 \mathcal{L}_{\text{state}} + \lambda_4\mathcal{L}_{\text{dist}}.
\label{eq:combined}
\end{align}
The $\{\lambda_i\}_{i=1}^4$ act as weighting hyperparameters.

Equation \ref{eq:reconstruction} encourages the KAE to  reconstruct states in the absence of any dynamics, promoting autoencoding. The linear prediction loss (Eq. \ref{eq:linear_prediction}) ensures that the observables evolve linearly in the latent space, while the state prediction loss (Eq. \ref{eq:state_prediction}) aids end-to-end prediction accuracy when mapping back to the state space. Finally, the encoder isometry (Eq. \ref{eq:isometry}) encourages preservation of inter-point distances even in the observable space. We discuss the significance of encoder isometry in Section \ref{sec:encoder_isometry}.

\subsection{Preprocessing Representations}
\label{sec:preprocessing}
Given we are working with neural representations, we draw from tools in RSA metrics literature. Permitting intra-layer comparison, these metrics first require embedding neural representations in a common space $\mathbb{R}^q$. Only then is a distance metric defined. \citet{lange23a} detail the intricacies and variations in this class of approaches. 

Our work is concerned solely with the initial embedding step. To avoid confusion with `embedding' in the context of Koopman approaches, we refer to this as \textit{preprocessing}. To elaborate, we apply the following preprocessing to $\mathbf{x_i}, \mathbf{x_j}$, before they are fed into a KAE:
\begin{align}
&\text{1. \textbf{Mean-centering}:} \ \mathbf{\hat{x}} = \mathbf{x} - \mathbb{E}[\mathbf{x}]
\\
&\text{2. \textbf{Projection}:}  \ \mathbf{\hat{x}} = \mathbf{\hat{x}}U_{:q}, \, \text{given } U\Sigma V^{\top}= \text{svd}(\mathbf{\hat{x}}) 
\\
&\text{3. \textbf{Normalizing}:} \  \mathbf{\hat{x}} = \mathbf{\hat{x}}/\|\mathbf{\hat{x}}\| 
\\
\begin{split}
&\text{4. \textbf{Procrustes alignment}:} \ \mathbf{\hat{x}} = \mathbf{\hat{x}}R, \\
&\qquad\text{where } R \in \mathcal{O}(q) \text{ solves } \min_R \|\mathbf{\hat{x}} - \mathbf{\hat{y}}R\|_F
\end{split}
\end{align}
Overall, we shift, project, and scale the representations before finding the best (rotational) alignment, making the representations more suited for comparison. In addition to affording us invariance properties, the preprocessing allows for learning a KAE on neural representations with originally non-uniform dimensions; i.e., outputs of differently-sized NN layers. However, we do not include models with non-uniform dimensions in our experiments. 

\subsection{Parametrization}
\label{sec:parameterization}
We parameterize the Koopman operator as
\begin{align}
\mathcal{K} = \exp{(\mathcal{G}/k)}^k, 
\end{align}
where $\mathcal{G}$ is another linear operator of the same shape and $k$ determines the number of steps that $\mathbf{\hat{x}}_i$ is advanced in observable space. When coupled with dimensionality reduction, this parameterization allows for a smooth $k$-step transformation of the neural activations, enabling an explicit visualization of topological changes. The parameterization is not restrictive: we can obtain the final prediction by directly applying the $k$-powered matrix.

Figure \ref{fig:hero} provides a visual summary of our methodology.

\section{Experiments}

We work with two residual MLPs, trained on the Yin-Yang \cite{kriener2022} and the MNIST classification tasks \cite{mnist}. Each of the MLPs consist of residual blocks (see Appendix \ref{app:dataset} for details). In all our experiments, we set $\mathbf{\hat{x}}_i$ as the first layer neural representations and $\mathbf{\hat{x}}_j$ as the penultimate layer representations of the residual MLP. Thus, when given $\mathbf{\hat{x}}_i$ as input, our KAEs are trained to predict $\mathbf{\hat{x}}_j$.

Given the parameterization described in Section \ref{sec:parameterization}, our KAEs can predict $k-1$ intermediate representations in observable space, before finally predicting $\mathbf{\hat{x}}_j$. Each of these observable space predictions can be decoded into state space via the KAE decoder for analysis. Ultimately, the output $\mathbf{\hat{x}}_j$ is fed into the final MLP layer, resulting in a class prediction. So, our KAEs can act as surrogate models, handling the intermediate computations. The classification accuracy provides a way to measure the surrogate quality of our KAE. Table \ref{table:accuracy} demonstrates that our KAEs are able to faithfully produce the penultimate layer representations for both datasets. We provide more details of the KAE architecture and their training in Appendix \ref{app:koopman}.

\subsection{Encoder Isometry}
\label{sec:encoder_isometry}
Typical implementations of KAEs \cite{takeishi2017, lusch2018deep, azencot2020forecasting, berman2023multifactor} do not consider encoder isometry. However, neural representations are topological objects; our isometry objective (Eq. \ref{eq:isometry}) promotes the observables to carry over the original shape of the representation. 

\begin{figure}[H]
    \centering \includegraphics[width=\columnwidth]{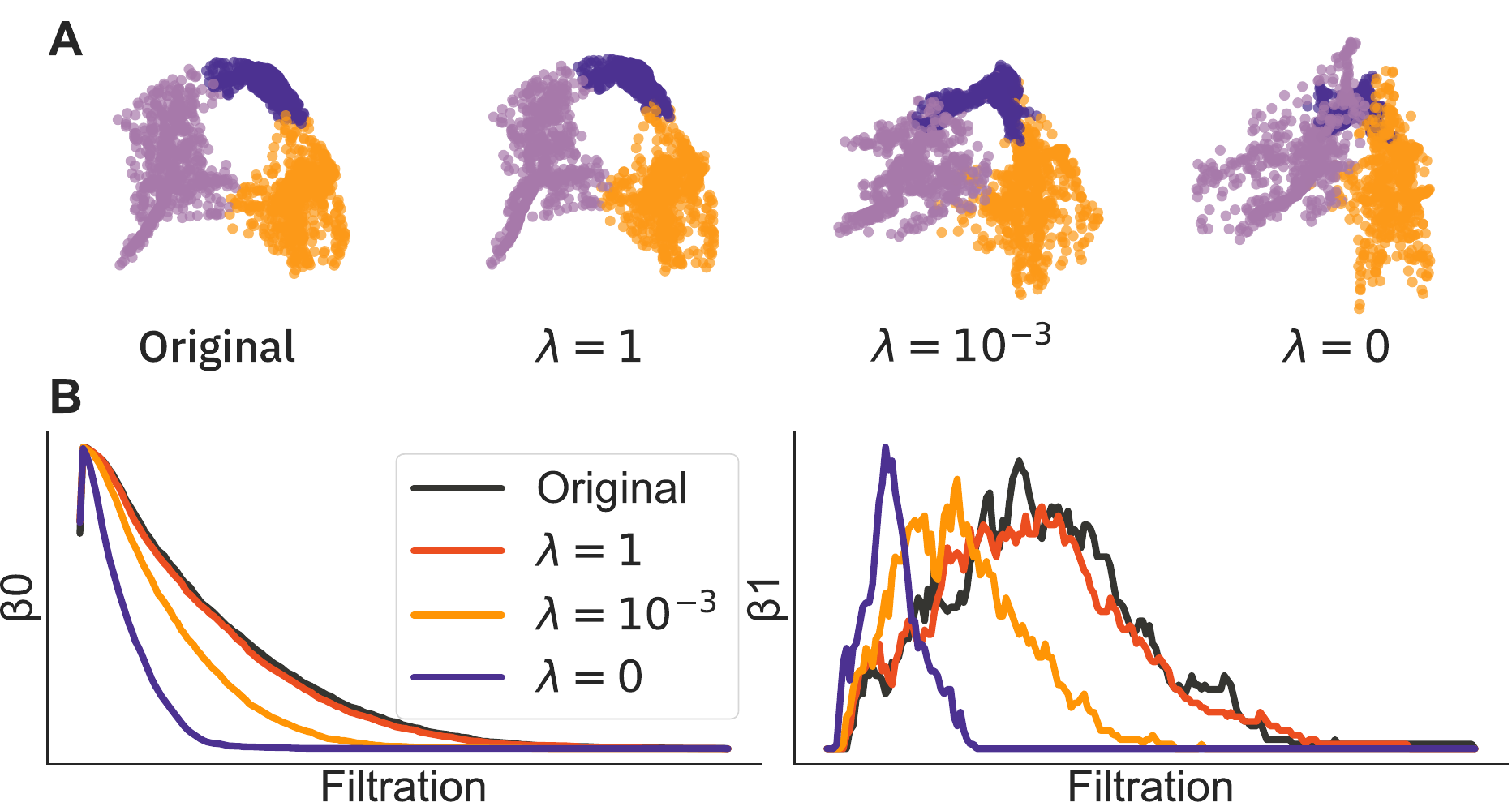}
    \vskip -0.1in
    \caption{(A) Each scatter plot displays $2\times10^3$ points projected onto the top three principal components (PCs) derived from representations in the penultimate layer. The leftmost plot shows PCs from the original MLP representations, while the remaining show PCs computed after embedding the representations into observable space via different KAEs. All PCs are aligned via the orthogonal Procrustes problem. (B) Betti curves, for $\beta0 \text{ and } \beta1$, across a filtration threshold of $\epsilon=4$ for the penultimate layer representations of the original model (black) and the observable space representations via different KAEs.}
    \label{fig:isometry_plot}
\vskip -0.2in
\end{figure}



\begin{table*}[ht!]
\caption{Caption}
\label{table:accuracy}
\vskip 0.1in
\begin{center}
\begin{small}
\begin{sc}
\begin{tabular}{lcclr}
\toprule
Dataset & MLP Top-1 \% Acc. & KAE Top-1 \% Acc. (StDev.) & Target Class & Edited Acc. (StDev.)\\
\midrule
\multirow{3}{*}{Yin-Yang} & \multirow{3}{*}{99.31} & \multirow{3}{*}{98.75 (0.15)} & Class 0 (Yin) & 98.78 (1.18) $\rightarrow$ 85.01 (1.90) \\
 &  &  & Class 1 (Yang) & 98.27 (0.21) $\rightarrow$ 78.88 (8.53) \\
 &  &  & Class 2 (Dots) & 99.97 (0.05) $\rightarrow$ 62.52 (1.35) \\
\midrule
\multirow{3}{*}{MNIST} & \multirow{3}{*}{99.03} & \multirow{3}{*}{98.53 (0.04)} & Class 1 & 99.23 (0.04) $\rightarrow$ 0.0 (0.0) \\
 &  &  & Class 4 & 98.29 (0.08) $\rightarrow$ 0.0 (0.0) \\
  &  &  & Class 7 & 98.01 (0.18) $\rightarrow$ 0.0 (0.0) \\
\bottomrule
\end{tabular}
\end{sc}
\end{small}
\end{center}
\vskip -0.1in
\end{table*}

To demonstrate, we train 3 KAE variants with different penalization strengths ($\lambda_4 =\{0, 10^{-3}, 1\}$) on the encoder isometry objective. The KAEs are trained to predict (and reconstruct) the penultimate layer representations of a residual MLP. Figure \ref{fig:isometry_plot}A displays the top three principal components of the penultimate layer representations in observable space. Figure \ref{fig:isometry_plot}B presents the \textit{Betti curves} of these same models, demonstrating that the most strongly penalized encoder (red) exhibits the closest topological similarity to the original model (black). These results indicate that increasing $\lambda_4$ leads to more topologically faithful representations in observable space. As a result, we expect that topological edits in the observable space will also be reflected in the state space.

\subsection{Simplifying Topology}

Given the parameterization described in Section \ref{sec:parameterization}, our KAEs can interpolate between $\mathbf{x}_i$ and $\mathbf{x}_j$ to produce intermediate representations. Remarkably, we demonstrate that the dynamics within our observable space naturally produce intermediate representations similar to those from the original MLP. To support this claim, we decode the observables into state space and quantify their topology. In Figure \ref{fig:topology}A, on the left, we present the $\beta_0$ Betti numbers of the neural representations from each block of a residual MLP trained to classify MNIST. As established in \citet{naitzat2020}, and evidenced by our plot, successive network layers generate increasingly simple topologies. In comparison, we also plot the $\beta_0$ Betti numbers of the decoded, intermediate outputs of five KAEs. Despite having no knowledge of the MLP's intermediate representations and their topologies, our KAEs still naturally simplify in topology at every step. As a visual aid, Figure \ref{fig:topology}B plots the top three principal components of selected iterations from one of the KAEs. 

The dynamics learnt by the KAEs produce a trajectory of neural representations with sound topologies, in line with what is found within a residual MLP. When paired with dimensionality reduction techniques, they provide an approximate visualization of how data is being transformed within a neural network. We hypothesize that the KAE dynamics can be made more faithful to the original residual network by regularizing the KAE's intermediate representations; for example, the KAE could be trained to predict all the neural representations from a residual network.

\subsection{Application: Model Editing}

The penultimate layer representations of well-trained classification models experience neural collapse (NC) \cite{papyan2020}, effectively `clustering' outputs, as seen at the bottom of Figure \ref{fig:topology}. In our case, the encoder isometry helps preserve this NC topology in observable space. As a result, identifying a class of `undesired' outputs in the penultimate layer is a straightforward task. Further, the dynamics that generate the outputs in observable space are governed by a linear operator. Hence, finding the undesired inputs, corresponding to the unwanted outputs, is a matter of applying the inverse operator $\mathcal{K}^{-1}$. To summarize, in observable space, we can quickly identify the unwanted outputs in a neural representation (due to NC) along with their corresponding inputs (by applying the inverse linear operator). Then, with the aid of a model editing algorithm, such as EMMET \cite{gupta2024}, we can learn an edited linear operator which generates an updated representation---sans the unwanted outputs. If the edited linear operator can maintain the rest of the topology, we can unlearn a specific class without affecting the model's performance on the other classes. We elaborate on our methodology in Appendix \ref{app:editing}.

\begin{figure}[ht!]
  \centering \includegraphics[width=0.49\textwidth]{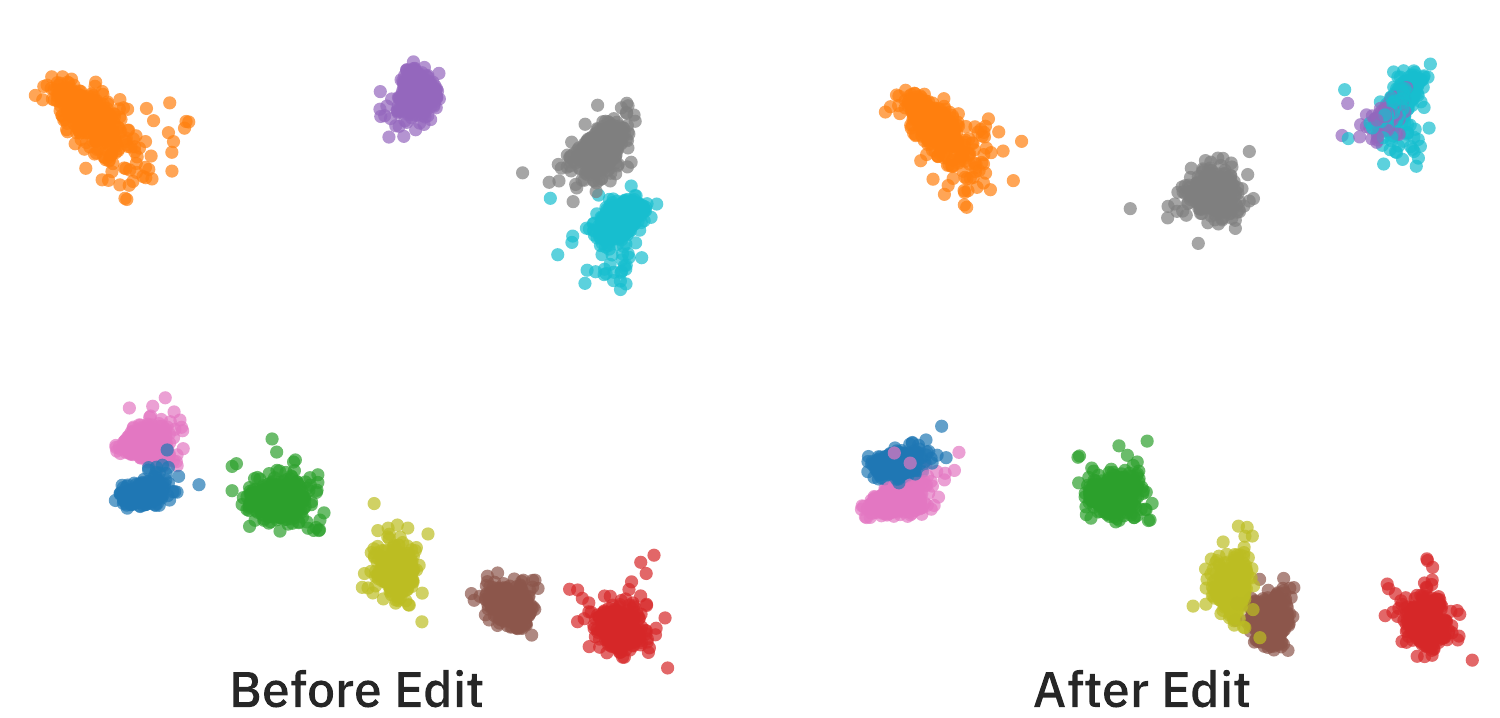}
  \vskip -0.1in
  \caption{$10^4$ points projected on the top three principal components of the neural representations produced by the Koopman operator in observable space before editing (left) and after editing (right). The KAE is trained on the first and penultimate-layer representations of a MNIST classifier. The operator is edited to forget class 4 (violet) by merging the outputs of that class with those of class 9 (light blue). The result of the merge is visible on the top right corner.}
  \label{fig:edited}
\vskip -0.1in
\end{figure}

Table \ref{table:accuracy} reports our model editing efforts for two datasets, with starkly different results, highlighting the importance of the neural collapse property. For the Yin-Yang dataset, we use the most strongly regularized KAE (see Figure \ref{fig:isometry_plot}). Despite performing sufficient class separation, the neural representation of the original MLP (and the KAEs), do not exhibit neural collapse; there is a large within-class variance in the penultimate layer. On the other hand, the representations of the MLP (and our KAEs) trained on MNIST exhibit strong neural collapse (see Figure \ref{fig:topology}). As a result, model editing is successful on the MNIST dataset but performs poorly on the Yin-Yang dataset. In Figure \ref{fig:edited}, we show the top three principal components of the penultimate representations before and after the linear operator is edited. Here, we edit the operator to remove class 4 (violet) by redirecting it to the class 9 (light blue) cluster, effectively merging the two classes. As a result, the KAE surrogate unlearns class 4. We found that the modified representations do not affect the performance of the KAE decoder and the subsequent MLP classifier on the remaining classes.


\section{Limitations and Future Work}

Tying together interpretability insights from the perspectives of topology and dynamical systems, our work introduces Koopman autoencoders as surrogate models, which learn the dynamics underlying a deep network's neural representations. By parameterizing the linear operator, we can interpolate an arbitrary number of steps between neural representations. And, our experiments validate that the generated interpolation follows the established principle of progressively simplifying topology. Additionally, we demonstrate how linear dynamics in observable space can enable editing the neural representations, leading to class unlearning. For future work, several  directions emerge:
\begin{itemize}[itemsep=1pt, topsep=0pt, partopsep=0pt]
    \item \textbf{Representation regularization:} Currently, our approach is limited to interpolating between two neural representations. How do we regularize the dynamics to interpolate through all the intermediate representations of a model?
        
    \item \textbf{Operator interpretability:} Given that a Koopman operator governs our dynamics, does spectral analysis of the operator offer insights into the original model's mechanism?

    \item \textbf{Observable space shaping:} Since we have the freedom to shape how neural representations look in observable space, are there other favorable topologies that enable certain goals (e.g., disentanglement, interpretability, unlearning)?

    \item \textbf{Architecture extensions:} Extending our approach to models with different architectures (e.g., convolutional layers, transformer blocks, etc.) could enable more sophisticated model editing applications beyond classification tasks. Can we extend our framework to unlearn concepts in language models?
    
\end{itemize}

In conclusion, our work demonstrates how Koopman theory can provide a practical framework for working with neural representations, opening new avenues for analyzing deep networks through the lens of dynamical systems.


\bibliography{ref}
\bibliographystyle{icml2025}

\newpage
\appendix
\onecolumn
\section{Dataset and model details}
\label{app:dataset}
\subsection{Yin-Yang task} 
The Yin-Yang dataset \cite{kriener2022} is a task with two-dimensional inputs consisting of three classes, allowing for easy visualization of the model's decision boundary and topology. For our experiments, we use a residual MLP architecture
$$
\text{Residual MLP}: \mathbb{R}^2 \rightarrow \text{Linear}(2 \rightarrow 10, \text{ReLU}) \rightarrow 4 \times [\text{ResBlock}(10, \text{ReLU})] \rightarrow \text{Linear}(10 \rightarrow 2)
$$

We generate a training dataset of $5\times 10^3$ samples, with roughly equal distribution among the three classes. For the test dataset, we generate another set of $5\times 10^3$ samples with a different seed. The network is trained to a test accuracy of $99.31\%$ using SGD with momentum (set to 0.9) for 500 epochs. We use a batch size of 512 samples, a weight decay set to $5\times10^{-4}$, and a cyclic learning rate peaking at $10^{-1}$.

\subsection{MNIST task}

For the MNIST task \cite{mnist}, we train a residual MLP with four blocks
$$
\text{Residual MLP}: \mathbb{R}^2 \rightarrow \text{Linear}(2 \rightarrow 784, \text{ReLU}) \rightarrow 4 \times [\text{ResBlock}(784, \text{ReLU})] \rightarrow \text{Linear}(784 \rightarrow 2)
$$
The model is trained to a test accuracy of $99.03\%$ using SGD with momentum (set to 0.9) for 30 epochs on a batch size of 128 samples, a weight decay set to $5\times10^{-4}$, and a cyclic learning rate peaking at $10^{-1}$.

\begin{figure}[h]
\begin{center}
    \centerline{\includegraphics[width=\columnwidth]{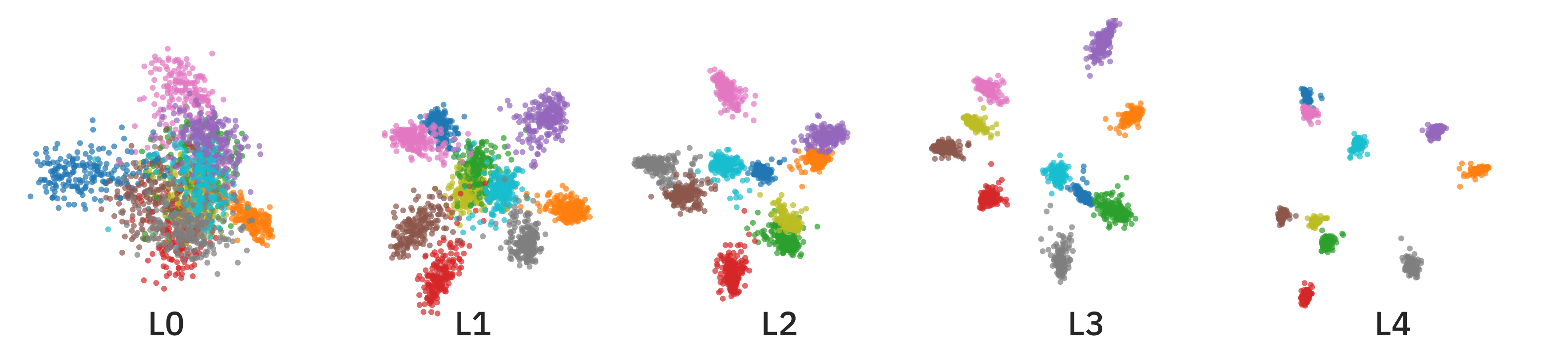}}
    \vskip -0.1in
    \caption{The top three principal components of the neural representations from the first layer (L0) and all residual blocks (L1-4) of a residual multi-layer perceptron (MLP). Each plot consists of $2\times10^3$ points and undergoes the preprocessing steps outlined in Section \ref{sec:preprocessing} before PCA. The model is trained on the MNIST digits task.}
    \label{fig:mnist_activations}
\end{center}
\vskip -0.3in
\end{figure}

Similar to Figure \ref{fig:yin_yang_activations}, we show the neural activations from each output layer of the MNIST model in Figure \ref{fig:mnist_activations}.

\section{Koopman autoencoder details}
\label{app:koopman}

We use the AdamW optimizer \cite{loshchilov2017decoupled} to train our KAEs. Table \ref{table:hparam} presents the hyperparameter choices and Table \ref{table:arch} outlines the architecture of the Koopman autoencoders used in both tasks.

\begin{table}[h]
\caption{KAE hyperparameter details}
\label{table:hparam}
\begin{center}
\begin{small}
\begin{tabular}{lccccccccc}
\toprule
Dataset & batch & observable dim. & \#epochs & $\lambda_\text{recon}$ & $\lambda_\text{linear}$ & $\lambda_\text{state}$ & $\lambda_\text{dist}$ & learning rate & weight decay \\
\midrule
Yin-Yang & 1024 & 20  & 1000 & 1 & 1 & 1 & 1 & $1\times10^{-1}$ &  $5\times10^{-4}$ \\
MNIST    & 512  & 800 & 100  & 1 & 1 & 1 & $10^{-3}$ & $5\times10^{-3}$ &  $5\times10^{-4}$ \\
\bottomrule
\end{tabular}
\end{small}
\end{center}
\end{table}

\begin{table}[h]
\caption{KAE architecture}
\label{table:arch}
\begin{center}
\begin{small}
\begin{tabular}{lll}
\toprule
\textbf{Component} & Yin-Yang & MNIST \\
\midrule
\multirow{3}{*}{Encoder} & batch $\times \mathbb{R}^{10}$ & batch $\times \mathbb{R}^{784}$ \\
 & Linear$(10 \rightarrow 30) \rightarrow$ LeakyReLU & Linear$(784 \rightarrow 1000) \rightarrow$ LeakyReLU\\
 & Linear$(30 \rightarrow 20)$ & Linear$(1000 \rightarrow 800)$ \\
\midrule
Koopman Matrix & batch $\times \mathbb{R}^{20}$ & batch $\times \mathbb{R}^{800}$ \\
 & Linear$(20 \rightarrow 20)$ & Linear$(800 \rightarrow 800)$ \\
\midrule
\multirow{3}{*}{Decoder} & batch $\times \mathbb{R}^{20}$ & batch $\times \mathbb{R}^{800}$ \\
 & Linear$(20 \rightarrow 30) \rightarrow$ LeakyReLU & Linear$(800 \rightarrow 1000) \rightarrow$ LeakyReLU \\
 & Linear$(30 \rightarrow 10)$ & Linear$(1000 \rightarrow 784)$ \\
\bottomrule
\end{tabular}
\end{small}
\end{center}
\end{table}

\newpage 
\section{Model Editing}
\label{app:editing}

We outline the steps of our model editing approach in Algorithm \ref{alg:editing}.

\begin{algorithm}[th]
   \caption{Model Editing with KAEs}
   \label{alg:editing}
\begin{algorithmic}
   \STATE {\bfseries Input:} trained KAE \{$\phi, \mathcal{K}, \phi^{-1}$\}, reprs. \{$\mathbf{x}_i, \mathbf{x}_j$\}, target class $c$
   \STATE {\bfseries Output:} Updated output reprs. $\mathbf{\hat{x}}_j$
   \STATE
   \STATE // Identify unwanted outputs
   \STATE $Z_{del} \leftarrow \{\phi(\mathbf{x}_j) \ | \ \mathbf{x}_j \text{ belongs to class } c\}$
   \STATE $Z_{keep} \leftarrow \{\phi(\mathbf{x}_j) \ | \ \mathbf{x}_j \text{ not in class } c\}$
   \STATE
   \STATE // Compute corresponding inputs
   \STATE $X_{mem} \leftarrow \mathcal{K}^{-1}\circ Z_{del}$
   \STATE $X_{keep} \leftarrow \{\mathbf{x}_i \ | \ \mathbf{x}_i \text{ not in } X_{mem}\}$
   \STATE
   \STATE // Select alternative outputs
   \STATE $Z_{new} \leftarrow \text{alt\_output}(X_c)$
   \STATE
   \STATE // Edit operator
   \STATE $\mathcal{L} \leftarrow \text{EMMET}(\mathcal{K}, \{X_{mem}, Z_{new}\}, \{X_{keep}, Z_{keep}\})$
   \STATE
   \STATE // Update reprs.
   \STATE $\mathbf{\hat{x}}_j \leftarrow \mathcal{L} \circ \mathbf{x}_i$
\end{algorithmic}
\end{algorithm}


\end{document}